\documentclass[lettersize,journal]{IEEEtran}
\usepackage{amsmath,amssymb,amsfonts}
\usepackage{algorithmic}
\usepackage[ruled]{algorithm2e}
\usepackage{array}
\usepackage{textcomp}
\usepackage{stfloats}
\usepackage{url}
\usepackage{color}
\usepackage{verbatim}
\usepackage{graphicx}
\usepackage{multirow}
\usepackage{array}
\usepackage{booktabs}
\usepackage{graphicx}
\usepackage{parskip}

\usepackage[numbers]{natbib}
\usepackage{bm}
\usepackage{mathtools}

\usepackage{subfigure}
\usepackage{makecell}
\usepackage{enumitem}

\usepackage{bibspacing}
\graphicspath{ {./fig/} }

\usepackage{stackengine}

\def\BibTeX{{\rm B\kern-.05em{\sc i\kern-.025em b}\kern-.08em
    T\kern-.1667em\lower.7ex\hbox{E}\kern-.125emX}}
\usepackage{balance}
\begin{document}

\title{Trust Semantics Distillation for Collaborator Selection via Memory-Augmented Agentic AI} 
\author{
Botao~Zhu,~\IEEEmembership{Member,~IEEE}, Jeslyn Wang, Dusit~Niyato,~\IEEEmembership{Fellow,~IEEE}, and Xianbin~Wang,~\IEEEmembership{Fellow,~IEEE}
\thanks{

B. Zhu and X. Wang (Corresponding author) are with Western University, Canada. J. Wang is with the University of Toronto, Canada. D. Niyato is with Nanyang Technological University, Singapore.
}
}

\maketitle

\begin{abstract}
 Offloading computational tasks from resource-constrained devices to resource-abundant peers constitutes a critical paradigm for collaborative computing. Within this context, accurate trust evaluation of potential collaborating devices is essential for the effective execution of complex computing tasks. This trust evaluation process involves collecting diverse trust-related information from every potential collaborator and performing trust inference based on the collected data. However, when each resource-constrained device independently assesses all potential collaborators, frequent data exchange and complex reasoning can incur significant overhead and further degrade the timeliness of trust evaluation. To overcome these challenges, we propose a task-specific trust semantics distillation (TSD) model based on a large AI model (LAM)-enabled teacher-student agent architecture. Specifically, the teacher agent is deployed on a server with powerful computational capabilities and an augmented memory module to perform multidimensional trust-related data collection, task-specific trust semantics extraction, and task-collaborator matching analysis. Upon receiving task-specific evaluation requests from device-side student agents, the teacher agent transfers the trust semantics of potential collaborators to the student agents, enabling rapid and accurate collaborator selection. Experimental results demonstrate that the proposed TSD model can reduce collaborator evaluation time, decrease device resource consumption, and improve the accuracy of collaborator selection.
\end{abstract}
 


\section{Introduction}

\IEEEPARstart{W}{ith} the growing complexity of computational applications over connected systems, individual resource-constrained devices become increasingly incompetent to execute complex computing tasks independently due to limited computational resources and capabilities. To address these limitations, a new collaborative computing paradigm, in which devices coordinate and share resources for collaborative task completion, has become a viable direction~\cite{9071995}. By leveraging distributed resources, this mechanism enhances the system's task completion capabilities and operational efficiency. Due to its advantages, distributed collaborative computing has been widely adopted across domains such as smart manufacturing, smart cities, and e-health, providing significant benefits including reduced latency and improved resource utilization~\cite{10508191}.

 To achieve effective task completion, the selection of reliable collaborators is a critical prerequisite. Previous studies have developed various approaches to evaluate collaborator reliability by analyzing factors such as historical behaviors, social attributes, and reputation metrics~\cite{10908235}. While these approaches provide valuable insights, they typically fail to capture the situational-dependent nature of reliability, particularly in terms of task-specific considerations. Recently, trust has emerged as a holistic criterion for collaborator evaluation and selection. In this context, trust is defined as the expectation of a resource-constrained task owner on a potential collaborator's capabilities, resources, and willingness for collaborative task completion~\cite{11096939}.

Within dynamic collaborative systems, the trustworthiness of a device is inherently time-varying due to the continuous influence of diverse factors, including task characteristics, collaboration environments, and the conditions of both task owners and collaborating devices. However, most existing trust evaluation techniques treat device trust as a static attribute with coarse labels such as trusted or untrusted, failing to effectively capture this dynamic nature of trust~\cite{10925363}. Moreover, these approaches consider trust as a generalized concept for all tasks, neglecting its task-specific nature. In practice, a collaborating device may display distinct levels of trustworthiness under different tasks. Therefore, a new mechanism is needed to accurately characterize the task-specific trustworthiness of collaborators within dynamic systems.

In this work, we introduce a new concept of trust semantics, inspired by the strengths of semantic representations in capturing contextual information, task-specific characteristics, and situational awareness~\cite{10589469}. Trust semantics models the trustworthiness of collaborating devices over time across multiple dimensions, including communication-related and computation-related, by evaluating their trust in each dimension and representing the evaluation results semantically. By extending the traditional binary notion of trust, trust semantics enables comprehensive and highly accurate evaluation of dynamic device trustworthiness, providing robust support for reliable collaborator selection. Nevertheless, due to the independent operation of devices and the inherent complexity of trust evaluation processes, assessing the trust semantics of collaborators by resource-constrained devices faces several challenges.

The first challenge stems from the overhead imposed on devices for collecting trust-related data. To accurately assess the trust semantics of its potential collaborators, a device must frequently collect data on their past collaborative behaviors, resources, and other relevant information~\cite{chain_of_trsut}. Frequent data exchange among devices generates a massive volume of data, which incurs a significant burden and processing delay. Moreover, due to limited memory capacity, individual devices are unable to store trust-related data for all potential collaborators. To mitigate this challenge, a practical solution is to deploy a server within the collaborative system as the trust evaluation center. By leveraging its abundant computational and storage resources, the server can collect, aggregate, and maintain trust-related data from all devices, thereby reducing device-side overhead.

The second challenge lies in accurately assessing the trust semantics of devices from the unstructured data collected with diverse quality. These data are gathered from a large number of devices and exhibit multi-source and heterogeneous characteristics, which necessitate complex processing operations, including consistency verification, feature alignment, and semantic fusion~\cite{10767305}. As a result, effective evaluation of trust semantics requires new mechanisms capable of leveraging heterogeneous information for precise semantic inference. Large AI models (LAMs), with their powerful semantic understanding, multimodal data fusion, and reasoning capabilities, are well-suited for extracting critical insights from complex data~\cite{10638533}, making them promising tools for trust semantics evaluation.

The third challenge arises from the continuous evolution of both the collaborative system environment and task-specific requirements. Device resources fluctuate over time due to changes in operational status and network conditions, which calls for dynamic trust evaluation mechanisms that can promptly capture such variations to ensure accurate assessment results~\cite{8897627}. Furthermore, task requirements may shift due to different objectives, necessitating adaptive trust evaluation for task-specific assessments of collaborators~\cite{10103199}. Owing to their capabilities in continuous learning and autonomous adaptation~\cite{11296817}, LAM-enabled agentic AI can effectively perceive environmental and task-level changes, making it suitable for enabling dynamic trust evaluation in evolving collaborative systems.

To address the above challenges, this study proposes a novel task-specific trust semantics distillation (TSD) model based on LAM-enabled agentic AI and a teacher-student agent architecture. The server-side teacher agent leverages comprehensive global data to extract task-specific trust semantics of potential collaborators for resource-constrained devices. Specifically, the distilled knowledge is transferred from the teacher agent to device-side student agents for lightweight collaborator selection. 
The main contributions of this paper are summarized as follows. 

    \textbullet \, We propose the new concept of trust semantics to characterize the temporal evolution of devices across multiple dimensions, providing a more accurate and task-specific assessment of device trustworthiness. Additionally, we utilize LAM-enabled agentic AI for accurate extraction of trust semantics from complex trust-related data.

    \textbullet \, We develop a new teacher-student agent architecture with a server-based trust evaluation. This architecture significantly reduces the communication and computational overhead on task owner devices for trust evaluation and collaborator selection.

    \textbullet \, We design an augmented memory module for the teacher agent to efficiently store, update, and retrieve task-specific trust semantics, enabling rapid collaborator selection.

\section{Agentic AI-Aided Teacher-Student Architecture for Trust Semantics Evaluation}

\begin{figure}[t!]
\centering
\includegraphics[scale=0.95]{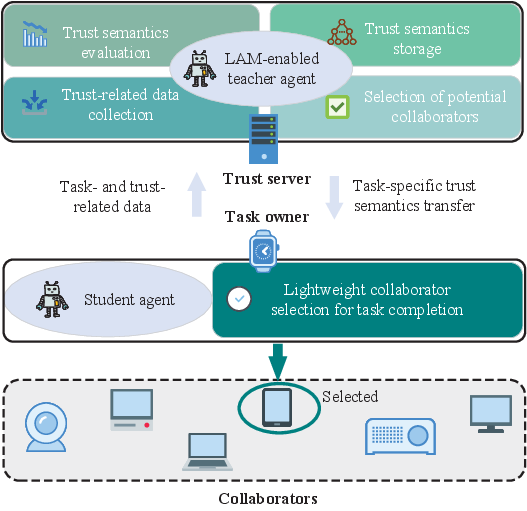}
\caption{A teacher-student architecture powered by LAM-enabled agentic AI for evaluating and transferring trust semantics between a trust server and task owners to enable task-specific collaborator selection.}
\label{system_model}
\end{figure}

LAM-driven agentic AI, endowed with autonomous intelligence, advanced reasoning, and goal-oriented execution, can effectively perform diverse tasks in complex environments. Individual agents with independent decision-making capabilities can form adaptive multi-agent systems, enabling cooperative task execution through coordinated interactions with other agents~\cite{9562559}. Building on these strengths, we propose the state-of-the-art TSD model, as shown in Fig.~\ref{system_model}. This model employs a teacher-student architecture powered by LAM-enabled agentic AI to perform trust semantic evaluation and knowledge transfer between a trust server (teacher agent) and resource-constrained task owners (student agents) for task-specific collaborator selection. The following analysis systematically examines the structure and advantages of the proposed model.

\subsection{LAM-Enabled Agentic AI for Trust Semantics Evaluation}

Benefiting from LAMs’ capabilities in multi-modal information integration, deep semantic representation, and complex reasoning~\cite{10558819}, LAM-enabled agentic AI could provide several key advantages for trust semantics evaluation.

\textbullet \, \textbf{Evolving trust evaluation via continuous interaction}. Given that the performance of collaborating devices can fluctuate, agentic AI can gather trust-related data through continuous interactions and learn from historical collaborations. This capability enables adaptive trust assessment that tracks changes in devices.

\textbullet \, \textbf{Task-specific trust evaluation via context-aware decision-making}. With inherent context-awareness, agentic AI can analyze collected trust-related data to identify task-dependent performance characteristics and behavioral patterns. Based on this analysis, it can infer task-specific trust semantics for devices, thereby supporting collaborator selection tailored to specific tasks.

\textbullet \, \textbf{Accurate trust semantic extraction}. Trust-related data are typically collected from diverse sources and span multiple dimensions. Leveraging the powerful multi-modal semantic understanding of LAMs, agentic AI can integrate, analyze, and reason over these heterogeneous data to extract precise trust semantics for collaborators.

\vspace{-0.1 in}
\subsection{Teacher-Student Agent Architecture}

Due to their limited computational resources, devices are not suitable for complex trust semantics extraction in dynamic environments. To address this, a server equipped with the LAM-enabled teacher agent is proposed to collect global trust-related data, extract trust semantics, and transfer the acquired knowledge to device-side student agents. This arrangement establishes a teacher-student agent architecture in which the server handles heavy computation and knowledge processing, while devices leverage the transferred knowledge to perform lightweight collaborator selection. The specific advantages of this architecture are outlined below.

\textbullet \, \textbf{Unified trust evaluation standard}. By possessing global trust-related data, the server-side agent can evaluate all devices using a unified assessment standard, thereby eliminating inconsistencies that may arise from local observation differences or independent evaluation by student agents.

\textbullet \, \textbf{Reduced communication and computation overhead}. By reporting trust-related data only to the server, devices avoid extensive inter-device data exchanges and complex local evaluations, resulting in a substantial reduction in communication and computational overhead.

\textbullet \, \textbf{Rapid collaborator selection}. 
Leveraging its computational capabilities and augmented memory module, the server-side teacher agent can rapidly respond to student agents' task requests by transferring task-specific trust semantics of potential collaborators. Student agents subsequently utilize the received information to select a final collaborator quickly.

\section{Task-Specific Trust Semantics Distillation}
This section presents the implementation of the TSD model, covering the system model, the evaluation and storage of task-specific trust semantics, and the transfer of trust semantics for collaborator selection.

\subsection{System Model}
We consider a collaborative system that includes a set of devices $\{a_1,\dots,a_I\}$ and a server. An LAM-enabled teacher agent is deployed on the server, while each device hosts an LAM-enabled student agent. These agents communicate by invoking the communication modules of their host devices. The server monitors all collaborations occurring within the system and records each collaborator's performance data. Every device can act as a task owner to generate computational tasks. We assume that the collaborative system supports three types of tasks: face recognition, video transcoding, and text word count~\cite{11096939}. For example, device $a_i$ initiates a face recognition task that contains a set of images and seeks to identify the number of people present in those images. The face recognition task can be described as \{``task type" : ``face recognition'', ``size'' : ``100 MB'', ``whether collaborator consent is required'' : ``yes'', ``minimum CPU requirement" :  ``2 GHz"\}, which specifies the storage and computing resources as well as the willingness to collaborate that potential collaborators must satisfy. All tasks are modeled with a common set of resource requirement dimensions, while the required values in each dimension may vary across tasks. The set of resource requirement dimensions is extensible. To select a suitable collaborator, the task owner $a_i$ evaluates potential collaborators from two aspects: historical collaboration trustworthiness and resource trustworthiness. Only collaborators considered trustworthy in both aspects are eligible for selection. In the following subsections, we provide a detailed description of the proposed TSD model, focusing on how trust evaluation of collaborators is conducted from these two aspects.

\begin{figure*}[t!]
\centering
\includegraphics[scale=1]{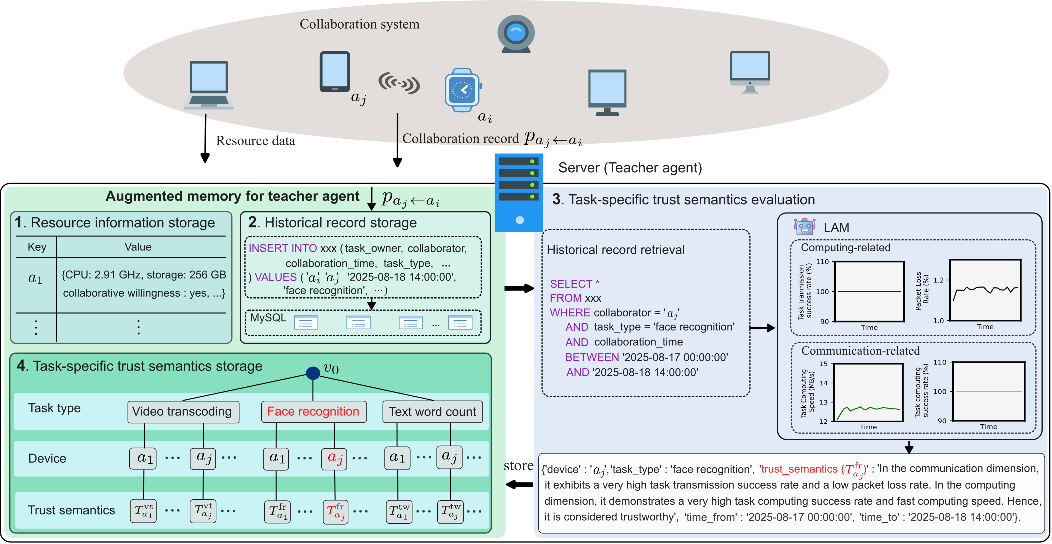}
\caption{Workflow of the server-side memory-augmented teacher agent for evaluating and storing task-specific trust semantics. 1) The teacher agent collects devices’ resource information and stores it in the resource information storage component. 2) The teacher agent collects collaboration data and stores the collected data in the historical record storage component. 3) The teacher agent evaluates devices’ task-specific trust semantics based on historical collaboration records. 4) The teacher agent stores the evaluated task-specific trust semantics in the trust semantics storage component.}
\label{teacher}
\end{figure*}

\vspace{-0.1 in}
\subsection{Task-Specific Trust Semantics Evaluation and Storage}

As the central authority for trust decision-making, the server-side teacher agent collects and stores trust-related data from devices in the system, including historical performance records and resource information. Based on these data, it extracts, updates, and maintains task-specific trust semantics for devices. Due to the LAM's limited internal memory, an external memory module is required to support storage and efficient retrieval. Therefore, we design the memory module with three distinct components: \textit{i) Resource information storage component} stores the resource information of all devices; \textit{ii) Historical record storage component} maintains historical performance data of all devices. \textit{iii) Trust semantics storage component} stores task-specific trust semantics of devices derived from historical performance data. Next, we present a detailed description of the teacher agent's workflow, highlighting how it uses the memory module.

\textbf{1) Resource information storage}. Devices report their idle-state resource information, including CPU, storage, collaborative willingness, and other relevant attributes, to the server. To maintain up-to-date records, devices promptly notify the server whenever these idle-state resources change. Note that resource changes occurring during task execution do not need to be synchronized with the server. The resource information storage component organizes each device’s resources in a key-value format, where the key identifies the device and the value encodes its resource attributes, as shown in Fig.~\ref{teacher}.

\textbf{2) Historical collaboration record storage}. The historical record storage component stores all historical collaboration records, ensuring that the teacher agent has sufficient knowledge regarding devices' historical performance. We consider a collaborative instance where collaborator $a_j$ helps the task owner $a_i$ complete a face recognition task. The server monitors this collaboration and collects the performance data of collaborator $a_j$, expressed as $p_{a_j \gets a_i}$ =  \{``task owner'' : ``$a_i$'', ``collaborator'' : ``$a_j$'', ``task type'' : ``face recognition'', ``collaboration time'' : ``2025-08-18 14:00:00'', ``task transmission success rate'' : ``100\%'', ``packet loss rate'' : ``1\%'', ``task computing speed'' : ``10 MB/second'', ``task computing success rate'' : ``100\%''\}. The task transmission success rate and packet loss rate characterize collaborator $a_j$’s communication-related performance, while the computing speed and computing success rate reflect collaborator $a_j$’s computing-related performance during the collaboration. In this paper, we consider only these four dimensions for all collaborators. It should be noted that additional performance dimensions can be incorporated if required. The server then stores $p_{a_j \gets a_i}$ in the historical record storage component of the memory module.
This component is implemented through a relational database, such as MySQL, where each record is stored in a structured format that includes task owner, collaborator, task type, collaboration timestamp, and associated communication and computation performance metrics.

\textbf{3) Task-specific trust semantics evaluation}. 
After storing a device’s historical performance data, the teacher agent promptly evaluates and updates its task-specific trust semantics. As shown in Fig.~\ref{teacher}, based on $p_{a_j \gets a_i}$, the teacher agent retrieves device $a_j$'s historical records for the face recognition task type within the recent time window--for example, from 2025-08-17 00:00:00 to 2025-08-18 14:00:00--from the historical record storage component. Note that the time span is flexibly configurable, enabling the extraction of device trust semantics over any specified period. The teacher agent then feeds these records into the LAM to extract the task-specific trust semantics. Because the LAM is initialized with communication- and computation-related knowledge, it can effectively examine whether a device’s communication and computing performance is trustworthy. In addition, to enhance accuracy, we employ few-shot learning, enabling the LAM to learn rules for evaluating trust semantics from a small number of samples. Based on the retrieved records, device $a_j$'s trust semantics for the face recognition task type is analyzed from four dimensions: task transmission success rate, packet loss rate, task computing success rate, and task computing speed. Finally, device $a_j$'s trust semantics is represented as \{``device'' : ``$a_j$'', ``task type'' : ``face recognition'', ``trust semantics ($T^{\text{fr}}_{a_j}$)'' : ``In the communication dimension, it exhibits a very high task transmission success rate and a low packet loss rate. In the computing dimension, it demonstrates a very high task computing success rate and fast computing speed. Therefore, it is considered trustworthy", ``time\_from" : ``2025-08-17 00:00:00", ``time\_to" : ``2025-08-18
14:00:00"\}.

\textbf{4) Task-specific trust semantics storage}. 
The trust semantics of each device is stored in the trust semantics storage component. To enable flexible management of trust semantics, we design a tree-structured storage module with three hierarchical layers: task type, device, and trust semantics. In this structure, all task types, devices, and task-specific trust semantics representations are represented as nodes, with the root node $v_0$ serving as the starting point without semantic meaning. Each node is expressed as a triplet $<$self, parent, child$>$, supporting structured storage and traversal operations. For instance, given the trust semantics extracted previously, the teacher agent first checks whether the face recognition node already exists in the tree. If absent, a new face recognition node is created. If present, the teacher agent then checks whether the device node $a_j$ is present under this task type. If absent, a new $a_j$ node is inserted along with its trust semantics as a child node. If node $a_j$ already exists, its trust semantics child node will be updated. Each device currently maintains a single trust semantics entry per task type, which can be extended to support multiple temporal entries for the same task type. The trust semantics storage component is implemented using a PostgreSQL database.

The teacher agent periodically repeats the above steps to keep the task-specific trust semantics representations of all devices up to date.

\subsection{Task-Specific Trust Semantics Transfer from Teacher to Student for Collaborator Selection}

When a resource-constrained task owner initiates a task, it sends a trust evaluation request to the server-side teacher agent. The teacher agent then queries the trust semantics storage component to identify suitable collaborators that meet the task’s requirements and returns their trust semantics to the task owner. Using the received information, the task owner can quickly select the final collaborator. For example, the task owner $a_i$ generates a video transcoding task  characterized by \{``task type" : ``video transcoding'', ``size'' : ``200 MB'', ``whether collaborator consent is required" : ``yes", ``minimum CPU requirement" : ``3 GHz"\}. It then sends the task request to the server, stating: ``I have a video transcoding task \{...\}, could you help me identify some trusted collaborators?''

\begin{figure}[t!]
\centering
\includegraphics[scale=0.98]{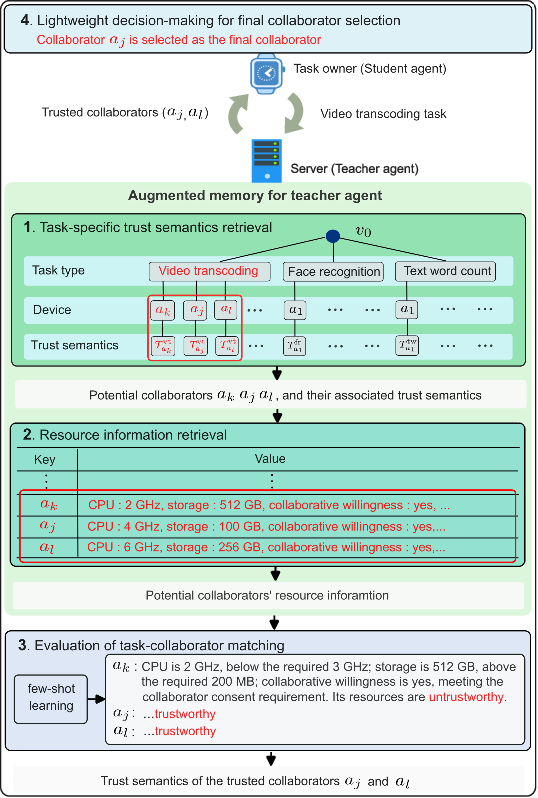}
\caption{Task-specific trust semantics transfer from the teacher agent to the student agent for collaborator selection. 1) The teacher agent retrieves the trust semantics of potential collaborators from the trust semantics storage component according to the task request initiated by the task owner. 2) The teacher agent retrieves the resource information of these potential collaborators from the resource information storage component. 3) The teacher agent performs task-resource matching analysis to assess resource trustworthiness and selects the trusted collaborators. 4) The student agent on the task owner selects the final collaborator using the trust semantics of trusted collaborators provided by the teacher agent.}
\label{teacher_student}
\end{figure}

\textbf{1) Task-specific trust semantics retrieval}. Upon receiving the task request, the teacher agent first uses the task type to locate the corresponding node in its tree-structured trust semantics storage component. It then scans all devices stored under that task-type node, identifies the trusted devices, and extracts their associated trust semantics representations. As shown in Fig.~\ref{teacher_student}, all trusted devices under the video transcoding node and their corresponding trust semantics are retrieved and represented as \{\{``device'' : ``$a_k$'', ``task type'' : ``video transcoding'', ``trust semantics ($T_{a_k}^{\text{vt}}$)'' : ``In the communication dimension, it exhibits a very high task transmission success rate and a low packet loss rate. In the computing dimension, it demonstrates a very high task computing success rate and fast computing speed. Therefore, it is considered trustworthy",...\}, \{``device'' : ``$a_j$'', ``task type" : ``video transcoding", ``trust semantics ($T^{\text{vt}}_{a_j}$)" : ``In the communication dimension, it exhibits a very high task transmission success rate and a low packet loss rate. In the computing dimension, it demonstrates a very high task computing success rate and fast computing speed. Therefore, it is considered trustworthy",...\}, \{``device'' : ``$a_l$'', ``task type" : ``video transcoding", ``trust semantics ($T^{\text{vt}}_{a_l}$)" : ``In the communication dimension, it exhibits a very high task transmission success rate and a low packet loss rate. In the computing dimension, it demonstrates a very high task computing success rate and fast computing speed. Therefore, it is considered trustworthy",...\}\}. As a result, devices $a_k$, $a_j$, and $a_l$ are identified as potential collaborators for the video transcoding task.

\textbf{2) Resource information retrieval}. 
 Since the trust semantics of these potential collaborators only reflect their historical collaboration trustworthiness, it is necessary to further evaluate whether their resources can satisfy the task’s requirements. Therefore, the teacher agent retrieves their resource information from the resource information storage component. As illustrated in Fig.~\ref{teacher_student}, the retrieved resource information of devices $a_k$, $a_j$, and $a_l$ is represented as \{``$a_k$" : \{``CPU" : ``2 GHz", ``storage" : ``512 GB", ``collaborative willingness" : ``yes",...\}, ``$a_j$" : \{``CPU" : ``4 GHz", ``storage" : ``100 GB", ``collaborative willingness" : ``yes",... \}, ``$a_l$" : \{``CPU" : ``6 GHz", ``storage" : ``256 GB", ``collaborative willingness" : ``yes",...\}\}.

\textbf{3) Evaluation of task-collaborator matching}. 
    Based on the resources of potential collaborators and the requirements of the task, the teacher agent evaluates whether their resources are trustworthy--namely, whether they can meet the task requirements. To support this process, the few-shot learning approach is adopted. By leveraging past task-collaborator pairs as examples, the LAM learns the underlying matching patterns and then applies the learned patterns to the current task to assess the resource trustworthiness of potential collaborators. This few-shot capability enables the teacher agent to produce accurate and consistent matching decisions even in rapidly changing environments, enhancing both adaptability and decision reliability. Therefore, based on the video transcoding task from the task owner $a_i$, the resource evaluation results for the potential collaborators are described as \{``$a_k$" : ``CPU is 2 GHz, below the required 3 GHz; storage is 512 GB, above the required 200 MB; collaborative willingness is yes, meeting the collaborator consent requirement. Therefore, its resources are untrustworthy.", ``$a_j$" : ``CPU is 4 GHz, above the required 3 GHz; storage is 100 GB, above the required 200 MB; collaborative willingness is yes, meeting the collaborator consent requirement. Therefore, its resources are trustworthy.", ``$a_l$" : ``CPU is 6 GHz, above the required 3 GHz; storage is 256 GB, above the required 200 MB; collaborative willingness is yes, meeting the collaborator consent requirement. Therefore, its resources are trustworthy."\}. According to the evaluation results, the potential collaborators $a_j$ and $a_l$ meet the requirements of the video transcoding task and thus are identified as the trusted collaborators. Then, the teacher agent delivers the trust semantics of the trusted collaborators to the task owner, including task-specific trust semantics derived from historical collaborations and resource-based trust evaluation results.

\textbf{4) Lightweight decision-making for final collaborator selection}. Based on the teacher agent's information, the student agent on the task owner selects the final collaborator via a lightweight decision process aligned with its objectives and preferences. As shown in Fig.~\ref{teacher_student}, device $a_j$ is selected as the final collaborator to execute the video transcoding task.

The proposed TSD model effectively reduces communication and computational overhead on devices while improving the accuracy, timeliness, and adaptability of task-specific trust evaluation, thereby providing a robust foundation for collaborator selection in dynamic environments.



\section{Experimental Analysis}

\begin{figure}[t!]
\centering
\includegraphics[scale=0.85]{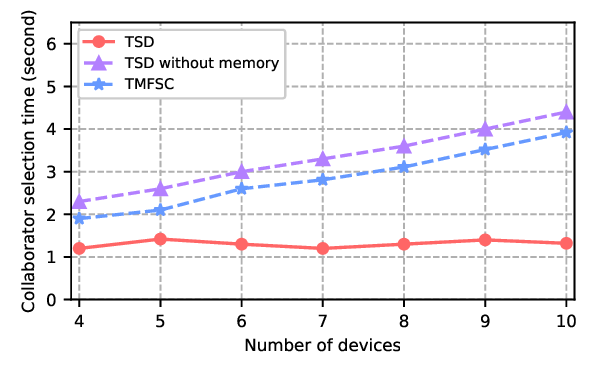}
\caption{By utilizing the trust semantics stored in the memory module of the teacher agent, the proposed TSD achieves faster collaborator selection than baseline methods.}
\label{spend_time}
\end{figure}

\begin{figure}[t!]
\centering
\includegraphics[scale=0.85]{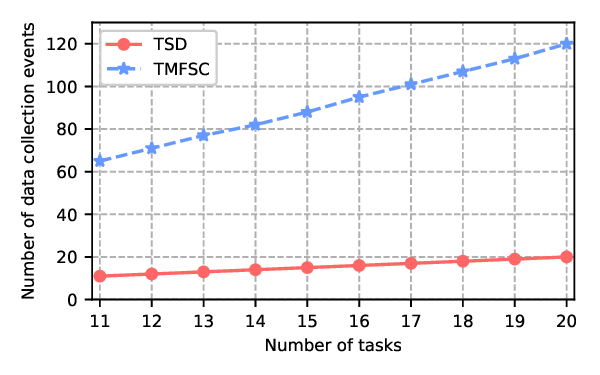}
\caption{The proposed TSD requires fewer data collection events than TMFSC, as its teacher agent leverages the augmented memory module to store collaborators’ historical performance records and trust semantics.}
\label{collection_count}
\end{figure}

To validate the proposed approach, we use a set of MacBook, DELL 5280, and DELL EMC 5200 devices to form a collaborative system. All agents are implemented using the OpenAI Agents SDK. Agents interact with OpenAI's LAMs via API calls, while devices communicate over Wi-Fi.

Fig.~\ref{spend_time} compares the collaborator selection time, defined as the interval from task generation by the task owner to the completion of collaborator selection. As the number of devices increases, the collaborator selection time for the trust model with fitness-based clustering (TMFSC) \cite{10944812} and for TSD without the memory module both increase.  This is because these methods require additional time for trust-related data collection and trust evaluation. The TSD without a memory module takes slightly more time than TMFSC due to the inference overhead of the LAM. In contrast, the collaborator selection time of the proposed TSD method remains almost constant, as it only needs to extract stored task-specific trust semantics from the memory module and perform the task-collaborator matching analysis, thus significantly accelerating collaborator selection.

Fig.~\ref{collection_count} compares the number of data collection events, defined as the historical performance data collection occurrences across all devices in the system. The TSD model requires fewer data collection events than TMFSC, as its teacher agent leverages the augmented memory module to store collaborators’ historical performance records and trust semantics. Consequently, in each task, the proposed TSD utilizes the stored trust semantics for collaborator selection, eliminating the need for repeated data collection from potential collaborators. This significantly reduces communication and computational overhead on the devices. By contrast, TMFSC lacks a memory module and must gather performance data from all potential collaborators for each task, resulting in higher overhead.

\begin{figure}[t!]
\centering
\includegraphics[scale=0.85]{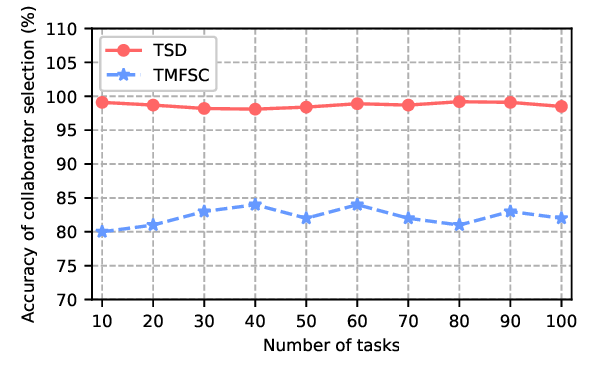}
\caption{The proposed TSD improves the accuracy of collaborator selection compared with TMFSC due to precise trust semantics extraction and task-collaborator matching analysis.}
\label{accuracy}
\end{figure}

Fig.~\ref{accuracy} compares the accuracy of collaborator selection. It can be observed that TSD significantly outperforms TMFSC, which can be attributed to two factors. First, the server in TSD maintains historical performance data for all devices, enabling the extraction of more accurate task-specific trust semantics. Second, leveraging LAM for task-collaborator matching analysis enhances matching precision, thereby improving the accuracy of collaborator selection.

\section{Future Directions}
The concept of trust semantics presents many promising research directions, with numerous opportunities for further exploration alongside the TSD model, particularly regarding their applications in realistic environments.

\textbullet \, \textbf{Evaluation and Mitigation of Environmental Changes on Dynamic Trust}. Future connected systems could face extremely dynamic operational environments. This is mainly due to dynamics in computing and communication resources as well as operational environments. By leveraging an LAM’s capabilities in environmental perception and semantic understanding, the influence of environmental factors on dynamic trust can be characterized more precisely.


\textbullet \, \textbf{Predictive Trust Distillation}. Given the complexity and inevitable processing delays, trust semantics evaluation and exchange in collaborative systems could take a significant amount of time--leading to reduced relevancy of trust evaluation due to evolving task requirements. To mitigate the undesired impact, predictive trust evaluation mechanisms could be incorporated into TSD for anticipated potential trust semantics, thereby enhancing the foresight and adaptability of decision-making on collaborator selection.


\textbullet \, \textbf{Trust Semantics Extraction Under Data-Missing Scenarios}. The current communication infrastructure experiences lots of difficulty in achieving deterministic and guaranteed data exchange. As a result, packet and data losses are unavoidable between the teacher and student agents in the proposed trust semantics distillation process. Given that trust-related data may be incomplete or missing in practice, accurately extracting trust semantics under realistic conditions of data sparsity or missingness constitutes an important research direction.

\section{Conclusion}
In this paper, we have proposed an innovative agentic AI-enabled teacher-student architecture for collaborator selection in dynamic systems. 
The server-side teacher agent is equipped with a memory module that leverages its powerful computational and storage capabilities to collect and store historical collaboration data while extracting and maintaining devices’ trust semantics.
Upon receiving a trust evaluation request from a resource-constrained task owner, the teacher agent transfers the trust semantics of potential collaborators to the student agent on the task owner, enabling it to perform lightweight collaborator selection locally. The proposed model significantly reduces computational and communication overhead on devices, facilitates rapid decision-making, and enhances the accuracy of collaborator selection by leveraging task-specific trust semantics. Furthermore, its scalable and robust architecture ensures reliability and adaptability in complex dynamic environments, providing a solid technical foundation for intelligent collaborative decision-making.

\footnotesize


\begin{thebibliography}{10}
\providecommand{\url}[1]{#1}
\csname url@samestyle\endcsname
\providecommand{\newblock}{\relax}
\providecommand{\bibinfo}[2]{#2}
\providecommand{\BIBentrySTDinterwordspacing}{\spaceskip=0pt\relax}
\providecommand{\BIBentryALTinterwordstretchfactor}{4}
\providecommand{\BIBentryALTinterwordspacing}{\spaceskip=\fontdimen2\font plus
\BIBentryALTinterwordstretchfactor\fontdimen3\font minus \fontdimen4\font\relax}
\providecommand{\BIBforeignlanguage}[2]{{%
\expandafter\ifx\csname l@#1\endcsname\relax
\typeout{** WARNING: IEEEtran.bst: No hyphenation pattern has been}%
\typeout{** loaded for the language `#1'. Using the pattern for}%
\typeout{** the default language instead.}%
\else
\language=\csname l@#1\endcsname
\fi
#2}}
\providecommand{\BIBdecl}{\relax}
\BIBdecl

\bibitem{9071995}
M.~Tang, L.~Gao, and J.~Huang, ``Communication, computation, and caching resource sharing for the {I}nternet of {T}hings,'' \emph{{IEEE} Commun. Mag.}, vol.~58, no.~4, pp. 75--80, Apr. 2020.

\bibitem{10508191}
Y.~Wang, C.~Yang, S.~Lan, L.~Zhu, and Y.~Zhang, ``End-edge-cloud collaborative computing for deep learning: A comprehensive survey,'' \emph{{IEEE} Commun. Surveys Tuts.}, vol.~26, no.~4, pp. 2647--2683, Fourth quarter 2024.

\bibitem{10908235}
M.~M. Rashid, Y.~Xiang, M.~P. Uddin, J.~Tang, K.~Sood, and L.~Gao, ``Trustworthy and fair federated learning via reputation-based consensus and adaptive incentives,'' \emph{{IEEE} Trans. Inf. Forensics Security}, vol.~20, pp. 2868--2882, Feb. 2025.

\bibitem{11096939}
B.~Zhu and X.~Wang, ``Networked physical computing: A new paradigm for effective task completion via hypergraph aided trusted task-resource matching,'' \emph{{IEEE} Trans. Netw. Sci. Eng.}, Jul. 2025, {E}arly {A}ccess, doi: 10.1109/TNSE.2025.3592859.

\bibitem{10925363}
T.~Luo, J.~Wang, Z.~Yan, and E.~Gelenbe, ``Graph neural networks for trust evaluation: Criteria, state-of-the-art, and future directions,'' \emph{{IEEE} Netw.}, vol.~39, no.~4, pp. 37--46, 2025.

\bibitem{10589469}
Y.~E. Sagduyu, T.~Erpek, A.~Yener, and S.~Ulukus, ``Will {6G} be semantic communications? opportunities and challenges from task oriented and secure communications to integrated sensing,'' \emph{{IEEE} Netw.}, vol.~38, no.~6, pp. 72--80, 2024.

\bibitem{chain_of_trsut}
B.~Zhu, X.~Wang, L.~Zhang, and X.~S. Shen, ``Chain-of-trust: A progressive trust evaluation framework enabled by {G}enerative {AI},'' \emph{{IEEE} Netw.}, vol.~39, no.~5, pp. 44--50, Sep. 2025.

\bibitem{10767305}
K.~Fan, J.~Guo, R.~Li, Y.~Li, A.~Liu, J.~Tang, T.~Wang, M.~Dong, and H.~Song, ``{RMDF-CV}: A reliable multi-source data fusion scheme with cross validation for quality service construction in mobile crowd sensing,'' \emph{{IEEE} Trans. Serv. Comput.}, vol.~18, no.~1, pp. 399--413, Jan.-Feb. 2025.

\bibitem{10638533}
F.~Jiang, Y.~Peng, L.~Dong, K.~Wang, K.~Yang, C.~Pan, D.~Niyato, and O.~A. Dobre, ``Large language model enhanced multi-agent systems for {6G} communications,'' \emph{{IEEE} Wireless Commun.}, vol.~31, no.~6, pp. 48--55, Dec. 2024.

\bibitem{8897627}
I.~Butun, P.~Österberg, and H.~Song, ``Security of the {I}nternet of {T}hings: Vulnerabilities, attacks, and countermeasures,'' \emph{{IEEE} Commun. Surveys Tuts.}, vol.~22, no.~1, pp. 616--644, First quarter 2020.

\bibitem{10103199}
S.~Pratap, P.~Dass, and S.~Misra, ``Cotev: Trustworthy and cooperative task execution in {I}nternet of {V}ehicles,'' \emph{{IEEE} Trans. Mobile Comput.}, vol.~23, no.~4, pp. 2915--2926, Apr. 2024.

\bibitem{11296817}
B.~Zhu, X.~Wang, and D.~Niyato, ``Task-specific trust evaluation for multi-hop collaborator selection via {GNN}-aided distributed agentic {AI},'' \emph{{IEEE} J. Sel. Areas Commun.}, pp. 1--16, Dec. 2025, {E}arly {A}ccess, doi: 10.1109/JSAC.2025.3642235.

\bibitem{9562559}
M.~Chen, D.~Gündüz, K.~Huang, W.~Saad, M.~Bennis, A.~V. Feljan, and H.~V. Poor, ``Distributed learning in wireless networks: Recent progress and future challenges,'' \emph{{IEEE} J. Sel. Areas Commun.}, vol.~39, no.~12, pp. 3579--3605, Dec. 2021.

\bibitem{10558819}
F.~Jiang, Y.~Peng, L.~Dong, K.~Wang, K.~Yang, C.~Pan, and X.~You, ``Large {AI} model-based semantic communications,'' \emph{{IEEE} Wireless Commun.}, vol.~31, no.~3, pp. 68--75, Jun. 2024.

\bibitem{10944812}
J.~Gao, C.~Cheong, M.~Zhang, Y.~Cao, T.~Peng, and S.~Pervez, ``A trust model with fitness-based clustering scheme in {FANETs},'' in \emph{Proc. IEEE Int. Conf. Trust, Secur. Priv. Comput. Commun.}, 2024, pp. 978--985.

\end{thebibliography}
\end{document}